\documentclass[conference]{IEEEtran}
\IEEEoverridecommandlockouts
\usepackage{multirow}
\usepackage{makecell}
\usepackage{subcaption}
\usepackage{cite}
\usepackage{amsmath,amssymb,amsfonts}
\usepackage{algorithmic}
\usepackage{graphicx}
\usepackage{textcomp}
\usepackage{xcolor,colortbl}
\definecolor{gray(x11gray)}{rgb}{0.75, 0.75, 0.75}
\usepackage{booktabs}
\def\BibTeX{{\rm B\kern-.05em{\sc i\kern-.025em b}\kern-.08em
    T\kern-.1667em\lower.7ex\hbox{E}\kern-.125emX}}
\newcommand{\CLGC}{\texttt{SAC-GCL}}
\begin{document}

\title{\texttt{SAC-GCL}: Structure-Aware Compression for Graph Contrastive Learning}

\author{\IEEEauthorblockN{Muhammad Ifte khairul Islam}
\IEEEauthorblockA{\textit{Department of Computer Science} \\
\textit{Texas A\&M University- Texarkana}\\
Texarkana,Texas, USA \\
mislam1@tamut.edu}
\and
\IEEEauthorblockN{Tanvir Hossain}
\IEEEauthorblockA{\textit{Department Of Computer Science} \\
\textit{Georgia State University}\\
Atlanta, GA, USA \\
thossain5@student.gsu.edu}
\and
\IEEEauthorblockN{Esra Akbas}
\IEEEauthorblockA{\textit{Department Of Computer Science} \\
\textit{Georgia State University}\\
Atlanta, Georgia, USA \\
eakbas1@gsu.edu}
}

\maketitle

\begin{abstract}
Graph Contrastive Learning (GCL) has achieved remarkable success in self-supervised representation learning by maximizing agreement between augmented views of a graph. Although augmented view generation is the central step in GCL, existing models rely on stochastic augmentation of the input graph, such as edge perturbation and node masking, which lack structural awareness and frequently destroy critical topology or introduce semantically inconsistent views. While recent adaptive methods improve upon random perturbations, they primarily operate on localized node or edge neighborhoods and fail to capture higher-order, macro-level structural organization crucial for graph classification. To address this limitation, we propose  \textbf{S}tructure-\textbf{a}ware \textbf{C}ompression for \textbf{G}raph \textbf{C}ontrastive \textbf{L}earning (\CLGC), a novel framework that employs structural decomposition and graph compression as a principled view generation strategy. Moving beyond random augmentations, \CLGC\ leverages structural decomposition techniques to extract cohesive subgraphs and condense them into supernodes, constructing coarsened graph views that preserve local density and capture higher-level structural patterns while reducing noise and redundancy. We leverage multiple cohesive subgraph models to derive complementary views of the graph, each capturing distinct structural properties. To capture multi-resolution information, a hierarchical view encoder processes these compressed views by initializing super-node features with shared-weight cohesive subgraph embeddings. Our results from extensive experiments on six benchmark datasets demonstrate that structure-aware compression provides a powerful alternative to random perturbations, yielding more robust and discriminative graph representations for graph classification tasks.\\
\end{abstract}


\section{Introduction}
Recently, graph contrastive learning (GCL) has emerged as a powerful self-supervised paradigm for graph representation learning, demonstrating state-of-the-art performance across both node-level and graph-level downstream tasks. GCL aims to maximize the agreement between representations of semantically consistent views of the same graph while distinguishing them from representations of other graphs~\cite{sun2019infograph,You2020GraphCL,icml2020_1971,zhu2020deep}. Recent models typically generate multiple graph views through stochastic augmentation strategies like random node dropping~\cite{You2020GraphCL}, edge deletion~\cite{velickovic2018deep}, and attribute masking~\cite{icml2020_1971}, implicitly assuming that semantic information is preserved under stochastic corruption. While these augmentations provide diversity and encourage the model to learn invariances, they often lack structural awareness and may disrupt crucial structural information or introduce semantically inconsistent views that inject noise into the contrastive learning objective.

This limitation has motivated the need for more structure-aware augmentation in graph contrastive learning~\cite{ju2025gps,du2025graph}. These methods aim to generate more informative views by formulating augmentation as a learnable process, leveraging model feedback or extracting local subgraphs. Although these adaptive schemes offer significant improvements over purely stochastic baselines by preserving local node centralities, they remain bound to local topological operations (e.g., node- or edge-level modifications). Consequently, they lack an explicit mechanism to capture, preserve, and leverage the higher-level structural organization inherent in complex networks. This drawback is particularly pronounced in graph-level classification tasks, where global topology, macro-scale connectivity, and multi-scale substructures dictate overall graph functionality and semantic identity.

On the other hand, graphs are organized into structurally dense regions that encode important semantic and functional information~\cite{liu2018graph}. In particular, cohesive subgraphs such as k-cores and k-trusses~\cite{akbas2017truss,khaouid2015k}  isolate densely connected and semantically meaningful regions that are robust to perturbations, making them well-suited for constructing stable and informative graph views. Unlike local ego-nets or isolated subgraphs, these cohesive structures capture higher-order topological interactions and retain the fundamental structural backbone of the network.

In this work, we propose \textbf{S}tructure-\textbf{a}ware \textbf{C}ompression for \textbf{G}raph \textbf{C}ontrastive \textbf{L}earning (\CLGC), a novel framework for graph-level representation learning. Unlike random augmentations or locality-preserving
subgraph sampling,  \CLGC\ introduces a structural decomposition and graph compression pipeline that explicitly preserves fine-grained connectivity while abstracting macro-level graph topology. Specifically, our framework extracts dense cohesive subgraphs and compresses them into higher-order super-nodes, constructing coarsened graph views that capture both internal local density and external inter-community relationships. By abstracting dense regions into super-nodes, these compressed views provide multi-scale structural representations that effectively filter out local topological noise and redundancy without losing fundamental connectivity patterns.

While our framework is generic and can accommodate various cohesive subgraph extraction algorithms, in this work we utilize $k$-core and $k$-truss decompositions due to their efficiency and complementary properties. Specifically, k-core captures node-level connectivity by enforcing minimum degree constraints, whereas k-truss captures edge-level cohesion through triangle-based reinforcement. Together, they enable the model to represent complementary aspects of structural organization in a scalable manner.

To learn effective representations, proposed structure-preserving compressed views are encoded hierarchically with dedicated view encoders and optimized through a joint objective that combines supervised classification with contrastive consistency across views.  
In the view encoder, we first generate subgraph embeddings via shared-weights GNN applied across all extracted cohesive subgraphs. These embeddings are then utilized as the initial features for the corresponding super-nodes in the compressed graph while learning the compressed graph embeddings. By jointly encoding both original and compressed views, the framework captures complementary structural information across resolutions. Overall, \CLGC~ produces robust, structure-aware representations that overcome the limitations of random augmentation-based methods. Our main contributions are summarized as follows:

\begin{itemize}
    \item We propose a novel contrastive learning framework for graphs that employs graph compression and structural decomposition as a deterministic, structure-aware view generation strategy,  moving beyond random perturbation-based augmentations.  
    \item We introduce a hierarchical encoding approach combining local (cohesive subgraphs) and global (super-node aggregation) information. With a joint objective combining supervised and contrastive loss, our model generates robust, structure-aware graph representations.
    \item We perform extensive experiments on benchmark graph classification datasets, demonstrating that \CLGC\ consistently outperforms state-of-the-art baselines while providing enhanced robustness against topological noise. 
\end{itemize}

\section{Preliminaries and Related Work}

In this section, we begin by introducing the preliminaries
and problem formulation. We then give the related work about GNN and graph contrastive learning.
\subsection{Preliminaries and Problem Formulation}\label{prbf}

We denote a graph as $G(V, A, X)$ where $V$ is the node-set, $A\in{\mathbb{R^{N\times N}}}$ is the adjacency matrix, and $X\in \mathbb{R}^{N\times d}$ is the feature matrix with $d$ dimensional node features and $N$ is the number of nodes in the graph. We represent a graph collection as $(\mathcal{G},Y)$ where $\mathcal{G}=\{G_{0},G_{1}, ..., G_{n}\}$ with $G_i$'s are graphs and $Y$ is the set of the graph labels. In this paper, we work on the graph classification problem, where the goal is to learn a function $f: \mathcal{G} \rightarrow {Y}$ to predict graph labels with a graph neural network in an end-to-end way.\\

\subsubsection{\textbf{Graph Neural Network}}
Graph Neural Networks (GNNs) have been widely used for static graphs in tasks such as node classification, link prediction, and graph classification. Among them, the Graph Convolutional Network (GCN)~\cite{Kipf} is one of the most widely adopted architectures. A GCN is a multilayer neural network that processes graph data by combining the features of each node with those of its neighbors while propagating information through edges.\\

Given an input graph $G = (V, A, X)$, where $X \in \mathbb{R}^{N\times d}$ is the node feature matrix, we set the initial hidden representation as $H^{(0)} = X$. A GCN layer updates node representations as:
\begin{equation*}
H^{(l+1)} = \sigma\!\left(\tilde{D}^{-\frac{1}{2}} \tilde{A} \tilde{D}^{-\frac{1}{2}} H^{(l)} \Theta^{(l)} \right),
\label{equ:gcn}
\end{equation*}
where $\tilde{A} = A + I$ is the adjacency matrix with self-loops, $\tilde{D}$ is the degree matrix of $\tilde{A}$, $H^{(l)}$ denotes the feature matrix at the $l$-th layer, $\Theta^{(l)}$ are trainable parameters, and $\sigma$ is a nonlinear activation function. Intuitively, each GCN layer aggregates information from one-hop neighbors, so an $L$-layer GCN captures information up to $L$ hops away. In our framework, GCN is used as the default instantiation of the message-passing operator $\text{GNN}^{(l)}(\cdot)$ in both levels of the view encoder.

For graph classification, the GCN is typically trained with a cross-entropy loss that compares the predicted labels with the ground-truth labels.

\subsubsection{\textbf{Contrastive Learning}}
Contrastive learning (CL) has become one of the most popular approaches for unsupervised representation learning, which learns through comparisons among different samples. This comparison is typically conducted between positive pairs of ``similar'' inputs and negative pairs of ``dissimilar'' inputs. Contrastive learning in the graph domain aims to learn a node or graph representation for a given input graph by maximizing the similarity between the different views of the input graph in their latent space via contrastive loss. In general, there are three parts to graph contrastive learning: i) View construction, ii) View encoder, and iii) Contrastive loss.

\begin{itemize}
\item \textbf{View Construction:} A view is represented as graph data, denoted as $v_i = (A_{i}, X_{i})$, where $A_{i}\in{\mathbb{R^{N\times N}}}$ and $X_{i}\in \mathbb{R}^{N\times d}$. While it is common to get 2 views, there are also some models that create more than 2 views~\cite{chen2023attribute, fang2023gomic}. In practical terms, view augmentation approaches involve techniques like node dropping~\cite{zeng2021contrastive}, edge perturbation~\cite{zhu2020deep}, attribution masking~\cite{You2020GraphCL}, and subgraph sampling~\cite{zhang2024motif}.

\item \textbf{View Encoder:} The view encoder maps the node representation or graph representation from different views of input graphs. As encoders for learning representations from views, any GNN model can be used based on the problem and type of graphs.

\item \textbf{Contrastive loss:} After learning the representations for different views using the view encoder, the contrastive learning model is optimized by a contrastive loss. The objective of the contrastive loss function is to maximize the similarity among different view representations of the same graph and minimize the similarity for view representations of other graphs. In general, the contrastive loss can be described as follows:
\begin{equation}
L_{contra} = -\log \left( \frac{sim(Z_{v_i},Z_{v_j})}{\sum_{j'=1}^{N}sim(Z_{v_i},Z_{v_{j'}})} \right)
\label{equ:contraL}
\end{equation}
where the $sim(\cdot)$ function measures the similarity between two representations. We can use any similarity function, like cosine similarity and mutual information (MI) in the contrastive loss. $Z_{v_i}$ and $Z_{v_j}$ are the $i$th view and $j$th view representations of node $v$.
\end{itemize}

\subsection{Related Works}
Graph neural networks (GNNs) have become the dominant approach for graph representation learning due to their ability to capture both node attributes and structural dependencies. Early models such as Graph Convolutional Networks (GCN)~\cite{Kipf} and Graph Attention Networks (GAT)~\cite{velivckovic2017graph} demonstrated the effectiveness of message passing in aggregating neighborhood information. Subsequent advances, including pooling and spectral techniques, have improved their ability to capture higher-order dependencies, which have proven particularly effective for graph classification in domains such as chemistry, bioinformatics, and social networks. However, these models generally rely on large-scale labeled data, which can be scarce in practice. This limitation has motivated the development of self-supervised approaches.

To address labeled data scarcity, Graph contrastive learning (GCL) models learn node and graph representations by maximizing agreement between augmented views of a graph~\cite{ju2024towards, tan2023s2gae,wu2024graph,sun2019infograph,lin2025simplified,hossain2026graph}. They relies heavily on view generation to construct correlated, self-supervised contrastive pairs that force the encoder to capture invariant semantic representations~\cite{ju2024towards}. Initial models in GCL such as GRACE~\cite{zhu2020deep} and GraphCL~\cite{You2020GraphCL} apply random augmentations such as node droping, edge removal and attribute masking to generate diverse graph views. While it is common to get two views, some models also create more than two views \cite{chen2023attribute,fang2023gomic}. While computationally efficient, they apply uniform perturbations across the entire graph and these perturbations are largely structure-agnostica, which disrupt important connectivity patterns and unintentionally degrading the cohesive regions of the graph~\cite{qin2025aagcn}. 

To address this issue, CI-GCL~\cite{10.5555/3692070.3694008} introduces a community-invariant augmentation strategy that preserves high-level community structure during contrastive view generation. It further unifies topology and feature augmentation through a spectral constraint, improving robustness against different types of graph noise. Recent research also explores other structure-aware augmentation strategies, such as motif-driven or cohesive subgraph-based methods (e.g., MicroGraph~\cite{zhang2024motif}, MHGCL~\cite{li2025mhgcl}). These methods aim to retain community-level or motif-level structure, but they still generate augmented views through perturbation or heuristic augmentation. More recent paradigms utilize learnable view generators or data-driven optimization to discover optimal views automatically~\cite{10499878}. While effective at capturing complex graph characteristics, generative and optimization-driven pipelines introduce significant computational overhead, require delicate joint optimization schedules, and remain susceptible to mode collapse or unstable training dynamics.

Graph compression and sparsification have been extensively studied in graph mining as a way to reduce complexity while preserving essential structures. Early works primarily utilized structural pruning to lower computational bottlenecks and mitigate message-passing pathologies. For example, TADA~\cite{lai2024efficient} utilizes topology-aware graph sparsification to augment the graph for the oversmoothing problem in deep Graph Neural Networks (GCNs). In parallel, recent self-supervised learning frameworks incorporate graph pooling and coarsening algorithms to hierarchically condense graph topology, generating multiscale views that capture multi-granularity representation~\cite{ju2025gps,du2025graph}. 
Unlike random perturbation-based or purely locality-preserving subgraph views, our approach leverages structural decomposition and targeted compression as a unified, structure-aware view generation pipeline. By systematically decomposing the graph into fundamental structural primitives (e.g., spectral components or dense subgraphs) and compressing redundant topology, our method preserves both global connectivity and fine-grained neighborhood semantics.

\section{Methodology\label{sec:method}}
\begin{figure}[t]
    \centering{
    \includegraphics[width=.5\textwidth]{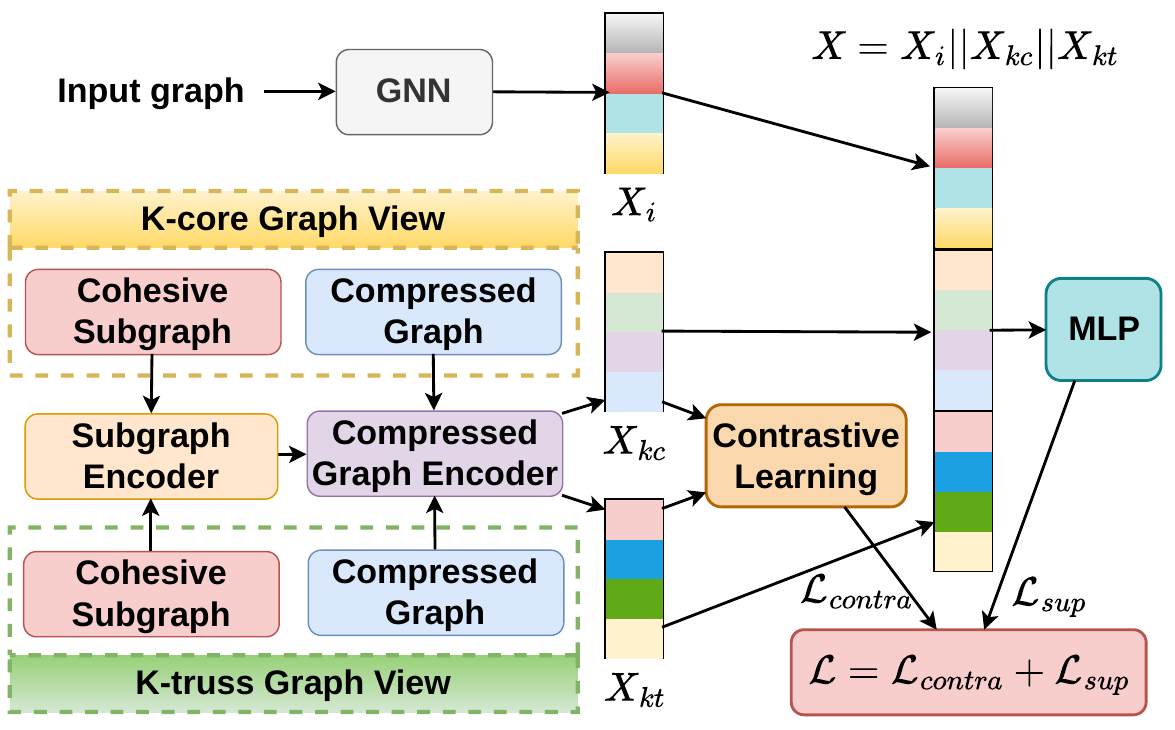}}
    \caption{Model Architecture: Graph Contrastive Learning with Structure-Aware Graph Compression}
    \label{fig:model}

\end{figure}

In this section, we present our proposed framework, Structure-aware Compression for Graph Contrastive Learning (\CLGC). \CLGC\ aims to enhance graph representations, particularly for graph classification tasks. The overall architecture is illustrated in Figure~\ref{fig:model} and comprises three main components: (i) \textbf{Graph View Generation via Cohesive Subgraph Compression}, (ii) \textbf{Structure-Aware View Encoder}, and (iii) \textbf{Contrastive Learning}.

\begin{figure*}[t]
    \centering{
    \includegraphics[width=.95\textwidth]{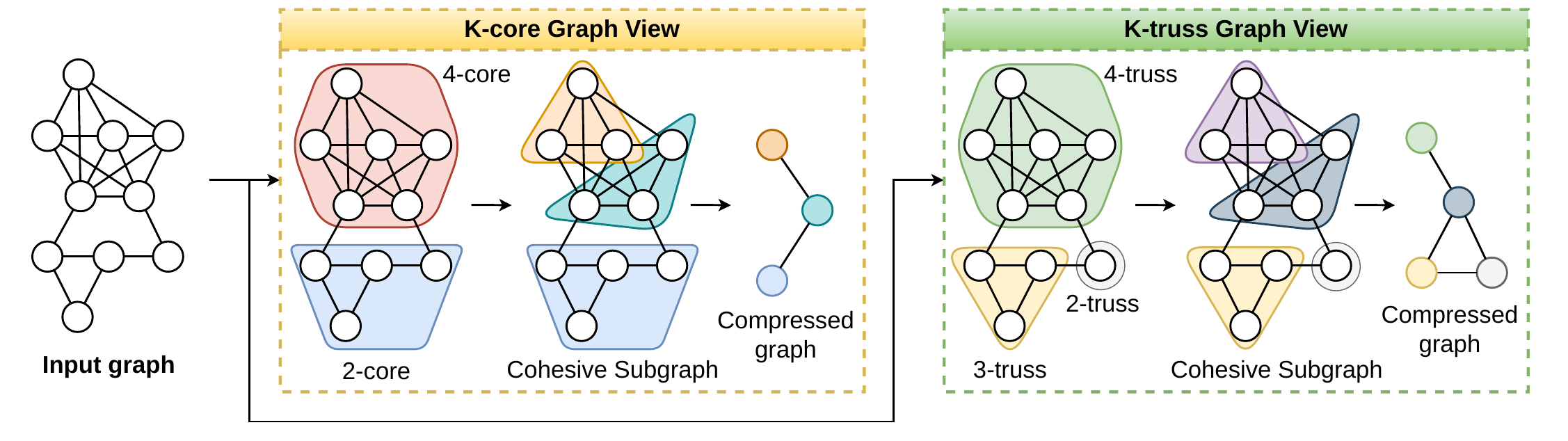}}
    \caption{Graph View Generation via Cohesive Subgraph Compression. Input graphs are decomposed into cohesive $k$-core and $k$-truss subgraphs with a size threshold. Extracted cohesive subgraphs are contracted into super-nodes to construct compressed graph views}. 
    \label{fig:view}
\end{figure*}
\subsection{\textbf{Graph View Generation via Cohesive Subgraph Compression}}
Unlike random augmentations, \CLGC\ leverages structural decomposition and compression methods to generate structurally distinct graph views from the original input graph $G$. Graph decomposition methods extract cohesive subgraphs that preserve fine-grained connectivity patterns. Figure~\ref{fig:view} illustrate the overview of graph view generation via cohesive subgraph compression.
While our framework is generic and can accommodate any method for extracting cohesive subgraphs, in this work, we utilize $k$-core and $k$-truss decompositions. These two decompositions capture fundamentally different notions of cohesiveness within a network: $k$-core identifies dense regions based on \emph{vertex degree}, while $k$-truss identifies cohesive structures based on \emph{triangle participation}. By leveraging both, we obtain complementary structural perspectives that enrich the subsequent contrastive learning process.\\

Next, we introduce a compression step that condenses cohesive subgraphs into supernodes, constructing compressed graphs as views that capture both local and global structural patterns. These compressed views, derived from cohesive subgraphs, provide higher order structural abstractions while preserving local density within supernodes.
\\~\\
{\textbf{$k$-core decomposition:}}
The $k$-core of a graph represents a subgraph where every node has a degree of at least $k$. Unlike node-level measures that quantify connectivity based on individual node degree, the k-core captures the collective connectivity of a group of nodes by identifying densely connected substructures within the graph. Formally, given an undirected graph $G = (V, E)$ and an integer $k \geq 1$, the $K$-core is the maximal induced subgraph $G_k = (V_k, E_k)$ such that:
\begin{equation*}
\forall v \in V_k, \quad \deg_{G_k}(v) \geq k.
\end{equation*}
Intuitively, the $k$-core decomposition iteratively removes nodes with degree less than $k$, along with their incident edges, until no such nodes remain. Higher values of $k$ produce more tightly connected subgraphs, representing the structural backbone of the network. The $k$-core hierarchy $G_k \subseteq G_{k-1} \subseteq \dots \subseteq G_1$ reveals connectivity layers of increasing cohesiveness. This property makes $k$-core decomposition a natural choice for view generation: it emphasizes progressively denser regions, providing an effective basis for forming compressed graph representations.\\

{\textbf{$k$-truss decomposition:}}
While $k$-core is based solely on the degree of nodes, $k$-truss considers \textit{triangle-based edge support}, capturing higher-order connectivity patterns in the graph. The support of an edge $e = (u, v) \in E$ in $G$ is defined as:
\begin{equation*}
sup_G(e) = \left| \{ w \in V : (u, w) \in E \ \wedge \ (v, w) \in E \} \right|,
\end{equation*}
which counts the number of triangles in which the edge participates. A subgraph $S = (V_S, E_S)$ is a \textbf{$k$-truss} if

$$\forall e \in E_S, \quad sup_G(e)\geq k - 2$$
where $k \geq 2$. The $k$-truss hierarchy is nested, i.e., $G_k \subseteq G_{k-1}$, and increasing $k$ yields progressively more cohesive triangle-rich regions. This property is particularly useful for contrastive view generation because triangle-rich subgraphs tend to correspond to community-like structures in real-world networks, which are structurally different from degree-based $k$-core regions.\\

{\textbf{Subgraph extraction and refinement:}}
For each decomposition method ($k$-core and $k$-truss), we compute subgraphs over a range of $k$ values, $k \in \{1,...,k_{max}\}$ for $k-$ core and $k \in \{2,...,k_{max}\}$ for $k$-truss, to capture structural diversity. However, a direct decomposition can yield overlapping subgraphs, as nodes or edges may appear in multiple $k$-levels. To avoid redundancy, we assign each node or edge to the highest-$k$ subgraph in which it appears, thereby preserving its most cohesive structural role. This strategy ensures that the extracted subgraphs are mutually exclusive and collectively cover the original graph’s cohesive regions. Specifically, each node $v$ is assigned a core number $k(v)$, defined as the largest $k$ for which $v$ belongs to the $k$-core of the graph. Nodes with the same core number are then grouped, and each connected component induced by these nodes defines a shell. Thus, a shell consists of a maximally connected set of nodes that share the same core number. We apply the same principle to $k$-truss decomposition. In both cases, each shell is treated as a cohesive subgraph, and the resulting shells are mutually exclusive while collectively representing the graph's cohesive structure.\\

While decomposition generates cohesive subgraphs, in practice, some subgraphs can be significantly large. Compressing these large subgraphs into a single compressed nodes may yield a very small compressed graph and thus information loss. To prevent excessive information loss within large cohesive regions, we impose a maximum size threshold $\lambda$ for any subgraph. If a cohesive subgraph exceeds this threshold, we split it into smaller connected subgraphs. To preserve the local connectivity of nodes within subgraphs, instead of a random split, we apply a Breadth-First Search (BFS) partitioning procedure. Specifically, for a large subgraph, we choose a random node as a seed, run BFS over the subgraph,  and split the subgraph into smaller connected components. The process repeats until all resulting subgraphs contain at most $\lambda$  nodes.  
This ensures that each super-node in the compressed graph 
represents a localized portion of the original community,
thereby retaining meaningful local structure while capturing global structure in a compressed graph.\\

{\textbf{Compressed graph construction:}}
We construct a compressed graph in which each subgraph is represented as a super-node. Nodes in a subgraph are combined into a super-node. This reduces the graph size by filtering low-level details while preserving high-level structural relationships between the cohesive components. Two super-nodes $i$ and $j$ are connected if any of their constituent nodes are adjacent in the original graph, and the edge weight is defined as
\begin{equation*}
A_{\text{super}}(i, j) = \frac{|E_{ij}|}{\min(|E_i|, |E_j|)},
\end{equation*}
where $E_{ij}$ denotes the set of edges between subgraphs $i$ and $j$, and $|E_i|$, $|E_j|$ are the number of edges internal to subgraphs $i$ and $j$, respectively. This normalization by the smaller of the two internal edge counts ensures that the connection strength between super-nodes reflects their relative interconnectivity rather than absolute size. The compressed graph preserves the connectivity between cohesive subgraphs while removing the fine-grained node-level details within each shell.\\

Compressed graphs enable the model to focus on relationships between higher-order structures (e.g., dense cores) rather than individual nodes, which can help reduce noise and improve representation quality of graphs. Also, by generating compressed views from two different decomposition strategies, we obtain complementary abstractions of the same underlying graph, which is critical.\\

Importantly, the $k$-core and $k$-truss decompositions capture different notions of structural cohesiveness and therefore produce different sets of cohesive subgraphs. Applying the above compression procedure independently to each decomposition yields two compressed graphs, denoted by ${G}^{c}{\mathrm{core}}$ and ${G}^{c}{\mathrm{truss}}$. Although both compressed graphs are derived from the same original graph ${G}$, their super-nodes and connectivity patterns differ because they are based on different structural properties. We therefore treat ${G}^{c}{\mathrm{core}}$ and ${G}^{c}{\mathrm{truss}}$ as two complementary structural views of the original graph. These views provide different but related representations of the graph's cohesive organization, which are subsequently used in a contrastive learning framework to encourage the model to learn representations that are consistent across the two structural perspectives while retaining view-specific structural information.

\subsection{\textbf{ Structure-Aware View Encoder}}
The view encoder generates informative embeddings from each compressed graph view using a two-level GNN strategy: \textit{subgraph-level encoding} and \textit{compressed-graph encoding}. The subgraph-level stage captures fine-grained local patterns within cohesive subgraphs, while the compressed-graph stage captures higher-order structural relationships between these subgraphs. We perform subgraph encoding first so that each cohesive subgraph is summarized into a meaningful vector representation, allowing the compressed-graph GNN to focus on inter-subgraph interactions rather than relearning local details, resulting in robust, structure-aware graph representations.\\

{\textbf{Subgraph-level Encoding}}
Let a cohesive subgraph be denoted as $G_s = (\mathcal{V}_s, \mathcal{E}_s, X_s)$, where $\mathcal{V}_s$ and $\mathcal{E}_s$ are its nodes and edges and $X_s$ is the initial features of the nodes in $G_s$. We apply a dedicated subgraph-level GNN (distinct for each view) with shared weights across all subgraphs within that view. The GNN updates node embeddings through message passing:
\begin{align*}
\mathbf{h}_v^{(l+1)} = \text{GNN}^{(l)}\Big(\mathbf{h}_v^{(l)}, \{\mathbf{h}_u^{(l)}: u \in \mathcal{N}_s(v)\}\Big),
\end{align*}

where $\mathbf{h}_v^{(0)} = \mathbf{x}_v$
After the final layer, we apply a readout function to obtain a single subgraph embedding:
\begin{equation*}
\mathbf{z}_s = \text{Readout}\Big(\{\mathbf{h}_v^{(L_s)}: v \in \mathcal{V}_s\}\Big),
\end{equation*}
 The embedding $\mathbf{z}_s$ captures the local structural patterns within each subgraph.\\  

{\textbf{Compressed Graph Encoding}}
 The subgraph embeddings $\mathbf{z}_s$ serve as the initial features for the super-nodes: as $\mathbf{h}_i^{(0)} = \mathbf{z}_{s_i}, \quad i \in \mathcal{V}_c.
$
We then apply a second GNN over the compressed graph:
\begin{align*}
\mathbf{h}_i^{(l+1)} = \text{GNN}^{(l)}\Big(\mathbf{h}_i^{(l)}, \{\mathbf{h}_j^{(l)}: j \in \mathcal{N}_c(i)\}\Big), \\ l = 0, \dots, L_c-1,
\end{align*}
where $\mathcal{N}_c(i)$ denotes the neighbors of super-node $i$ in the compressed graph, and $L_c$ is the number of layers.\\

This \textbf{two-level encoding} preserves both local subgraph information and global higher-order structure, resulting in \textbf{robust, structure-aware graph representations}.
In both subgraph and compressed graph encoding layers, as a readout function, we apply mean and max pooling and concatenate their outputs:
\begin{equation*}
Z = \frac{1}{N} \sum_{i=1}^N h_i \,\|\, \max_{1 \leq i \leq N} h_i,
\end{equation*}
where $h_i$ denotes the embedding of node $i$ and $\|$ represents concatenation. Mean pooling summarizes the global distribution of features, while max pooling captures salient structural signals. Applying this readout to the set of subgraph embeddings gives the subgraph-derived graph embedding $Z_{\text{sub}}$. Applying it to super-node embeddings gives the compressed-graph embedding $Z_{\text{comp}}$.
As the final embedding for a view, we concatenate these into one as $Z_{\text{final}} = Z_{\text{sub}} \,\|\, Z_{\text{comp}}$.\\

The same pipeline is applied independently to the $k$-core and $k$-truss compressed graphs, yielding the two final view embeddings $Z_{\text{core}}$ and $Z_{\text{truss}}$.

\subsection{ \textbf{Contrastive Learning}}
In the final stage, we align the two view representations while also optimizing for graph classification. We adopt a cross-view contrastive objective: for each graph, the embeddings from the $k$-core and $k$-truss views form a positive pair, while negatives are taken from \textit{all other graphs in the batch}, but always across views (i.e., $Z_{\text{core}}^{(i)}$ is contrasted against $Z_{\text{truss}}^{(j)}$ for $j \neq i$). This cross-view negative sampling, following GraphCL \cite{You2020GraphCL}, encourages the model to learn view-invariant representations across structurally different compressions. Instantiating Eq.~\eqref{equ:contraL} with cross-view positive pairs, the contrastive loss for graph $i$ is defined as:
\begin{equation}
\mathcal{L}_{\text{contra}}^{(i)} =
-\log \left(
\frac{\operatorname{sim}(Z_{\text{core}}^{(i)}, Z_{\text{truss}}^{(i)})}
{\sum_{j=1}^{N} \operatorname{sim}(Z_{\text{core}}^{(i)}, Z_{\text{truss}}^{(j)})}
\right),
\label{lcon}
\end{equation}
where $\operatorname{sim}(\cdot)$ is a similarity function such as cosine similarity, $Z_{\text{core}}^{(i)}$ and $Z_{\text{truss}}^{(i)}$ are the paired embeddings of the same graph from the two views, and $Z_{\text{truss}}^{(j)}$ represents embeddings of other graphs in the batch.




The embeddings from the two views, $Z_{\text{core}}$ and $Z_{\text{truss}}$, are concatenated to form a joint representation  as $ Z_{\text{joint}} = Z_{\text{core}} \,\|\, Z_{\text{truss}}$
and then  $Z_{\text{joint}}$ is fed into a multi-layer perceptron (MLP) for graph classification. As the supervised loss,   cross-entropy (CE) loss is used as
\begin{equation}
\mathcal{L}_{\text{sup}} = -\sum Y \log \hat{Y}, 
\quad \hat{Y} = \text{MLP}(Z_{\text{joint}}).
\label{lsup}
\end{equation}

The overall training objective combines supervised and contrastive losses given in Equations \ref{lcon} and \ref{lsup}:
\begin{equation*}
\mathcal{L} = \alpha \mathcal{L}_{\text{sup}} + (1 - \alpha) \mathcal{L}_{\text{contra}},
\end{equation*}
where $\alpha$ balances label supervision and cross-view consistency.



\subsection{Theoretical Insight on Compression-Based View Alignment}
\label{sec:theory}
Random perturbation–based view generation in graph contrastive learning 
often disrupts global connectivity by removing nodes or edges arbitrarily. 
In contrast, our compression-based strategy constructs structurally coherent 
graph views by deterministically grouping cohesive regions (via $k$-core and $k$-truss)
and representing their interconnections at a higher level of abstraction. 
Although intra-community edges are collapsed during compression, 
the local structural context is largely preserved through the subgraph 
splitting mechanism applied before compression.
Aligning multiple compressed views—each derived from a distinct
decomposition such as k-core and k-truss—encourages the model to learn rep-
representations that are invariant to redundant internal structure yet sensitive to
the graph’s global organization.\\ 

\paragraph{Mutual Information Perspective.}
Let $G=(V,E,X)$ denote the original graph and 
$f_c$ a compression mapping that produces a compressed view $G' = f_c(G)$.
The mutual information (MI) between $G$ and $G'$ is defined as:
\begin{equation*}
I(G;G') = H(G) - H(G|G'),
\end{equation*}
where $H(\cdot)$ denotes Shannon entropy. 
Compression differs from random perturbation because it removes redundant edges 
systematically rather than randomly, thus reducing entropy loss. 
When large dense regions are collapsed into super-nodes, the conditional entropy term 
$H(G|G')$ is dominated by information within the intra-subgraph structures:
\begin{equation*}
H(G|G') \leq H(G_{\text{intra}}),
\end{equation*}
where $G_{\text{intra}}$ denotes the set of edges and nodes internal to compressed subgraphs. 
Since these regions are highly cohesive and internally redundant, the lost information 
is relatively small compared with random edge deletions that destroy inter-community links.\\

\paragraph{Proposition (Informal).}
For a graph $G$ with modular or hierarchical organization, 
let $f_c^{(1)}$ and $f_c^{(2)}$ denote two complementary compression mappings 
(e.g., $k$-core and $k$-truss). 
If both mappings preserve inter-community connectivity 
and each subgraph is locally constrained by the size limit $\lambda$, then
\begin{equation*}
I(G_{c}^{(1)}; G_{c}^{(2)}) \geq I(G_{\text{pert}_1}; G_{\text{pert}_2}),
\end{equation*}
where $G_{\text{pert}_1}$ and $G_{\text{pert}_2}$ are randomly perturbed views.
Thus, compression-based contrastive learning retains more informative and 
topologically faithful relationships between graph views.
\section{Experiments}\label{sec:exp}

\begin{table}[b]
\small
\centering
\caption{Dataset statistics. $|G|$ is the number of graphs, $V_{avg}$ and $E_{avg}$ denote the average number of nodes and edges per graph, and $|C|$ is the number of classes.}
\label{table:table1}
\renewcommand{\arraystretch}{1.2}
\begin{tabular}{c c c c c }
\hline 
\textbf{Dataset} & $|G|$ & $V_{avg}$ & $E_{avg}$ & $|C|$ \\ \hline\hline
D\&D & 1178 & 284.32 & 715.66 & 2 \\
 PROTEIN & 1113 & 39.06 & 72.82 & 2 \\
 IMDB-BINARY & 1000 & 19.77 & 96.53 & 2 \\
 IMDB-MULTI & 1500 & 13.00 & 65.94 & 3 \\
 REDDIT-BINARY & 2000 & 429.63 & 497.75 & 2 \\
 COLLAB & 5000 & 74.49 & 2457.78 & 3 \\
 \hline
\end{tabular}
\end{table}

\subsection{\textbf{Datasets}} 
We use six publicly available datasets summarized in Table~\ref{table:table1}, all of which are commonly used for benchmarking graph classification models \cite{morris2020tudataset}. These datasets include four social network datasets (IMDB-BINARY, IMDB-MULTI, REDDIT-BINARY, and COLLAB) and two biological or chemical datasets (D\&D and PROTEIN). Each dataset differs in terms of graph size, number of nodes and edges, and class distribution, providing a diverse evaluation scenario.

\noindent The datasets vary significantly in size and domain. For instance, REDDIT-BINARY and COLLAB contain large graphs representing user interactions, while D\&D and PROTEIN contain more complex molecular structures with rich biological features. This diversity ensures a robust evaluation of our model across different types of graph data.

\begin{table*}
\centering
\small
\caption{ \centering{ Graph Classification Performance (Accuracy \% ). The best results are highlighted in \textbf{bold}, the second-best results are \underline{underlined}.}}
\label{table:accClust}
 \small
\begin{tabular}{|c|c |c c c c c c|}
 \hline

\textbf{Method} & \textbf{Model} &
\textbf{IMDB-B} &
\textbf{IMDB-M} &
\textbf{REDDIT-B} &
\textbf{COLLAB} &
\textbf{D\&D} &
\textbf{PROTEIN} \\
\hline\hline
  \multirow{6}{*}{\makecell{Supervised \\ Learning}} & gPool &73.40& 50.27 &74.70& 77.58 & 75.0 & 71.9\\
  & SAGPool &73.00& 49.43 & 84.66& 70.10& 75.7 & 72.5\\
  & MinCutPool & 70.78 & 52.8 & 75.67 & 69.91 & 76.7 & 76.3\\
  & DiffPool & 68.40 & 45.62 & 66.65 &74.83 & 66.9 & 68.2\\
  & ASAP & 72.74 & - & - &78.95 & 76.9 & 74.2\\
  & MPool & 74.20 & 50.26 & 84.10 &\underline{83.62} & \underline{81.2} & \underline{79.30}\\\hline
  \multirow{6}{*}{\makecell{Contrastive \\ Learning}} &InfoGraph & 71.34 & 47.93 & 82.50 &72.36 & 72.85 & 74.09\\
  & GraphCL & 71.48 & 48.11 & 89.53 &72.36 & 78.62 & 74.32\\
  & S2GAE   & 75.76 &\underline{51.79} & 87.83 & 81.02 & - & 76.37\\
  & CTAug   & \underline{76.60} & 51.12 & \textbf{92.28} & 81.72 & - & 74.10\\
  & CI-GCL   & 73.85 & - & \underline{90.80} & 74.40 & 79.63 & 76.50\\
  & \textbf{\textit{SAC-GCL}}& \textbf{78.81} & \textbf{52.88}  & 89.65& \textbf{86.70}& \textbf{83.43} & \textbf{81.42}\\\hline

\end{tabular}
\end{table*}

\subsection{\textbf{Baseline:}}
\subsubsection{Graph Pooling Methods}
We compare against several hierarchical graph pooling methods, including gPool~\cite{gao2019graph}, SAGPool~\cite{pmlr-v97-lee19c}, MinCutPool~\cite{bianchi2020spectral}, DiffPool~\cite{ying2018hierarchical}, ASAP~\cite{ranjan2020asap}, and MPool~\cite{islam2023mpool}. These methods learn graph-level representations by progressively reducing the graph size while preserving task-relevant structural information.
\begin{itemize}
\item \textit{gPool}: selects important nodes by projecting node features onto a learnable vector and retaining the top-ranked nodes at each pooling layer.
\item \textit{SAGPool}: computes node importance scores using graph convolution, allowing the pooling operation to consider both node features and graph topology.
\item \textit{MinCutPool}: learns a soft assignment of nodes to clusters by optimizing a normalized min-cut objective together with an orthogonality regularization term.
\item \textit{DiffPool}: uses a GNN-based assignment function to group nodes into clusters, forming hierarchical graph representations in an end-to-end manner.
\item \textit{ASAP}: forms local clusters using attention-based neighborhood scoring and retains the most informative clusters for graph-level representation learning.
\item \textit{MPool}: combines multiple pooling operations in parallel and integrates their outputs to capture complementary structural information.\\
\end{itemize}

\subsubsection{Graph Contrastive Learning Methods}
We also compare with representative graph contrastive and self-supervised learning methods, including InfoGraph~\cite{sun2019infograph}, GraphCL~\cite{You2020GraphCL}, S2GAE~\cite{tan2023s2gae} CTAug~\cite{wu2024graph}, and CI-GCL~\cite{10.5555/3692070.3694008}.
\begin{itemize}
\item \textit{InfoGraph}: maximizes the mutual information between graph-level representations and representations of local patches, such as nodes and substructures.
\item \textit{GraphCL}: generates two augmented views of each graph using operations such as node dropping, edge perturbation, attribute masking, and subgraph sampling, and then aligns their representations with a contrastive objective.
\item \textit{S2GAE}: follows a reconstruction-based self-supervised strategy, where part of the graph structure is masked and the model is trained to recover the missing edges.
\item \textit{CTAug}: generates augmentations with awareness of cohesive subgraph structures, making it closely related to our method. However, it still relies on stochastic augmentation rather than deterministic compression.
\item \textit{CI-GCL}: preserves high-level community structure during graph contrastive learning by using community-invariant topology and feature augmentation. Unlike CI-GCL, which learns augmentation strategies over the original graph, our method explicitly constructs compressed graph views through structural decomposition and supernode-based coarsening.

\end{itemize}


\subsection{\textbf{Experimental Setup}} 
For each dataset, we perform 10 independent random splits using different seed values. The data is divided into 80\% for training, 10\% for validation, and 10\% for testing. We implement our model using PyTorch and PyTorch Geometric. The model is optimized using the Adam optimizer \cite{kingma2014adam}. Node embedding dimensionality is fixed to 128 across all datasets.

We tune hyperparameters using grid search over the following ranges: learning rate \{1e-2, 5e-2, 1e-3, 5e-3, 1e-4, 5e-4\}, weight decay \{1e-2, 1e-3, 1e-4, 1e-5\}, and pooling ratio \{1/2, 1/4\}. Each model is trained for a maximum of 100K epochs, with early stopping triggered if the validation loss does not improve over 50 consecutive epochs.\\

For fair comparison with baseline methods, we use GCN as the message-passing backbone in the main experiments, as it is the most commonly adopted architecture in prior work. Unless otherwise specified, we set the subgraph size parameter to $\lambda=15$ and the contrastive loss weight to $\alpha=0.6$, as these values yield the best performance in our ablation studies.

\subsection{\textbf{Results}}
Table~\ref{table:accClust} presents the graph classification accuracy of \CLGC\ and the baseline methods on six benchmark datasets. Overall, \CLGC\ obtains the highest accuracy on five out of six datasets, including IMDB-BINARY, IMDB-MULTI, COLLAB, D\&D, and PROTEIN. These results indicate that structure-aware graph compression can provide effective contrastive views for learning graph-level representations.\\

Compared with supervised graph pooling methods, \CLGC\ achieves consistent improvements across the benchmark datasets. On IMDB-BINARY, \CLGC\ obtains 78.81\%, compared with 74.20\% from MPool. On COLLAB and PROTEIN, \CLGC\ achieves 86.70\% and 81.42\%, respectively, while MPool obtains 83.62\% and 79.30\%. These improvements suggest that the proposed compression strategy differs from conventional pooling methods. Rather than only reducing the number of nodes, \CLGC\ constructs compressed views from cohesive subgraphs, allowing the model to preserve local structural information while capturing relationships among higher-level graph components.\\

Compared with graph contrastive learning baselines, \CLGC\ also demonstrates competitive performance. On IMDB-BINARY and IMDB-MULTI, \CLGC\ achieves 78.81\% and 52.88\%, improving over CTAug by 2.20 and 1.76 percentage points, respectively. Since these datasets represent social interaction graphs, the results suggest that deterministic compressed views based on $k$-core and $k$-truss decompositions can capture useful structural patterns for graph classification. On COLLAB, \CLGC\ reaches 86.70\%, outperforming the contrastive baselines reported in the table, including S2GAE, CTAug, and CI-GCL. This result indicates that modeling interactions among compressed cohesive regions can be beneficial for larger interaction networks.\\

On the biological datasets D\&D and PROTEIN, \CLGC\ achieves 83.43\% and 81.42\%, respectively. These results are higher than the reported pooling and contrastive baselines, suggesting that structure-aware compression is also effective for biological graph classification. In particular, the result on PROTEIN indicates that preserving cohesive substructures before graph-level representation learning can help retain informative structural patterns that may be reduced by direct pooling or random augmentation.\\

For REDDIT-BINARY, CTAug obtains the highest accuracy with 92.28\%, while \CLGC\ achieves 89.65\%. Although \CLGC\ does not produce the best result on this dataset, it remains comparable to other contrastive baselines such as GraphCL and CI-GCL and slightly improves over GraphCL. This observation suggests that cohesive-subgraph-aware stochastic augmentation may be particularly effective for this dataset, while the proposed deterministic compression framework provides more consistent performance across different graph domains. Overall, the results show that combining complementary $k$-core and $k$-truss compressed views within a contrastive learning framework improves graph representation learning across a diverse set of datasets.





\subsection{Ablation Studies}
To better understand the contributions of different components and design choices in our framework, we conduct a series of ablation studies. Below we present the results for three aspects: component analysis, the effect of the contrastive weight, and the effect of subgraph size.

We compare the performance of the full model (\CLGC) with three ablated variants: \CLGC$_{sup}$ (trained with only supervised loss), \CLGC$_{core}$ (using only the $k$-core encoder), and \CLGC$_{truss}$ (using only the $k$-truss encoder) in Table~\ref{table:result2}. Across all datasets, the full model consistently outperforms its variants. The results highlight the complementary benefits of using both graph compression views ($k$-core and $k$-truss) as well as the importance of the contrastive objective. Removing either the compression view or the contrastive loss reduces performance, confirming that each component plays a critical role in the success of our framework.

\begin{table}[t]
\centering
\normalsize
\caption{ \centering{Performance comparisons with the main model ($\CLGC$), the model with supervised loss ($\CLGC_{sup}$), the model with $k$-core view encoder ($\CLGC_{core}$), and the model with $k$-truss view encoder ($\CLGC_{truss}$).} }
\label{table:result2}
\resizebox{\linewidth}{!}{
\begin{tabular}{|c|c |c |c|c|}
 \hline
{\cellcolor{gray!60} \textbf{Dataset}} & \rotatebox{50}{\textbf{\cellcolor{gray!60} $\textbf{\CLGC}$}} & \rotatebox{50}{\textbf{\cellcolor{gray!60} $\textbf{\CLGC}_{sup}$}} & \rotatebox{50}{\textbf{\cellcolor{gray!60} $\textbf{\CLGC}_{core}$}} & \rotatebox{50}{\textbf{\cellcolor{gray!60} $\textbf{\CLGC}_{truss}$}}\\\hline

 {IMDB-B }& \textbf{78.81} & 75.10  & 74.95 & 75.07   \\
{  IMDB-M }& \textbf{52.88} & 50.26  & 50.15  & 49.88 \\
{ REDDIT-B} & \textbf{89.65} & 88.20 & 86.32& 87.54\\
{ COLLAB} & \textbf{86.70} & 84.25  & 83.65  & 83.79  \\
{ D\&D }& \textbf{83.43} & 80.85  & 79.54  & 80.32   \\
{  PROTEIN} & \textbf{81.42}  & 77.63  & 75.77  & 75.56  \\
  \hline
  
\end{tabular} 
}
\end{table}

\begin{figure}[t]
    \centering
    \includegraphics[width=0.95\columnwidth]{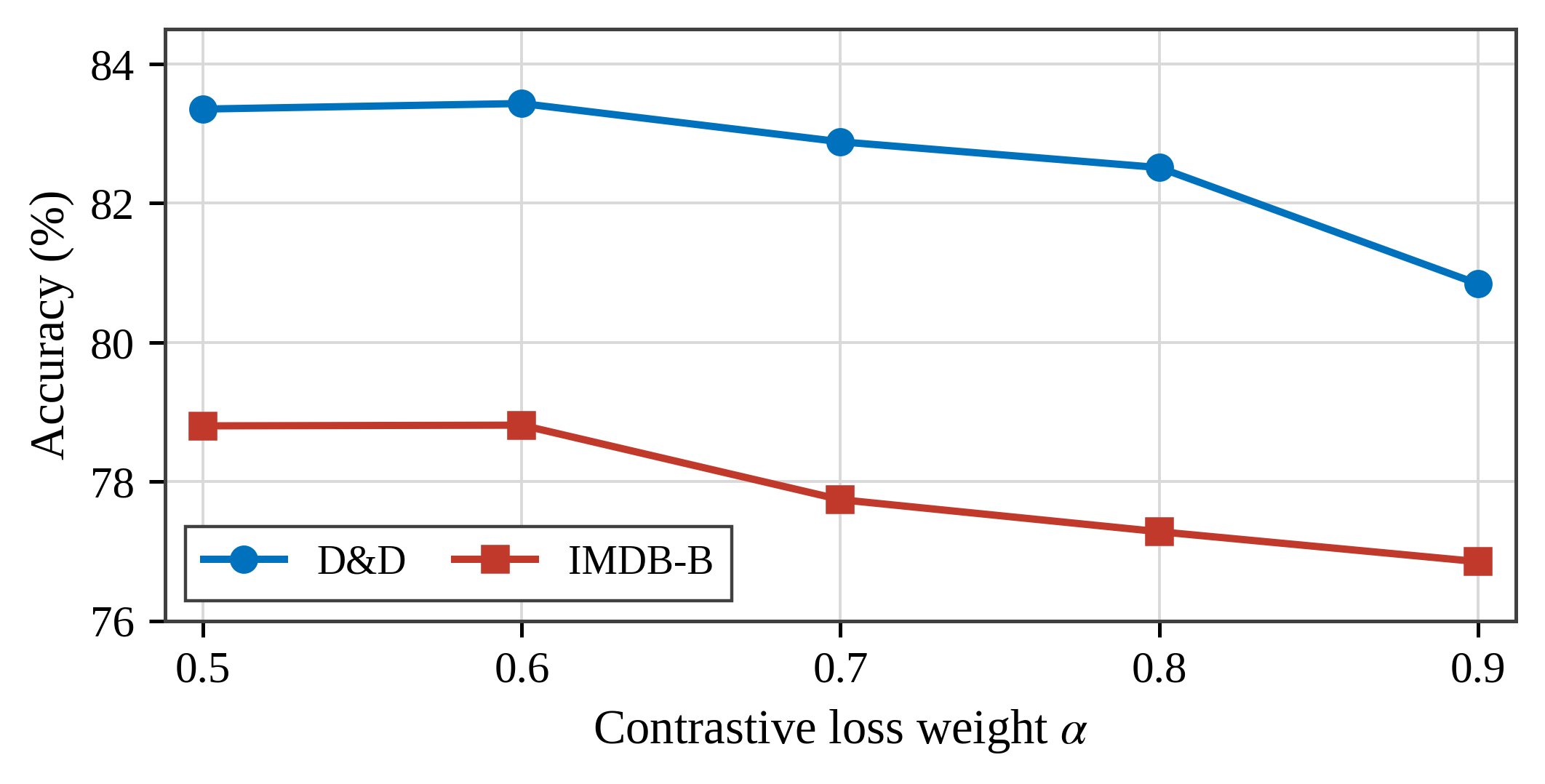}
    \caption{Performance of \CLGC\ for different values of the contrastive loss weight $\alpha$ on D\&D and IMDB-B.}
    \label{alpha}

\end{figure}

We further evaluate the influence of the loss weight parameter $\alpha$, which balances supervised and contrastive learning objectives, in Figure~\ref{alpha}. 
The results indicate that performance peaks when $\alpha=0.6$. Both smaller and larger values of $\alpha$ reduce accuracy, suggesting that an appropriate trade-off between the two objectives is critical for optimal representation learning. When the contrastive weight is too small, the model cannot fully benefit from the structural alignment between views; when it is too large, the model becomes dominated by contrastive learning and underutilizes task supervision.

\begin{figure}[t]
    \centering
    \includegraphics[width=0.95\columnwidth]{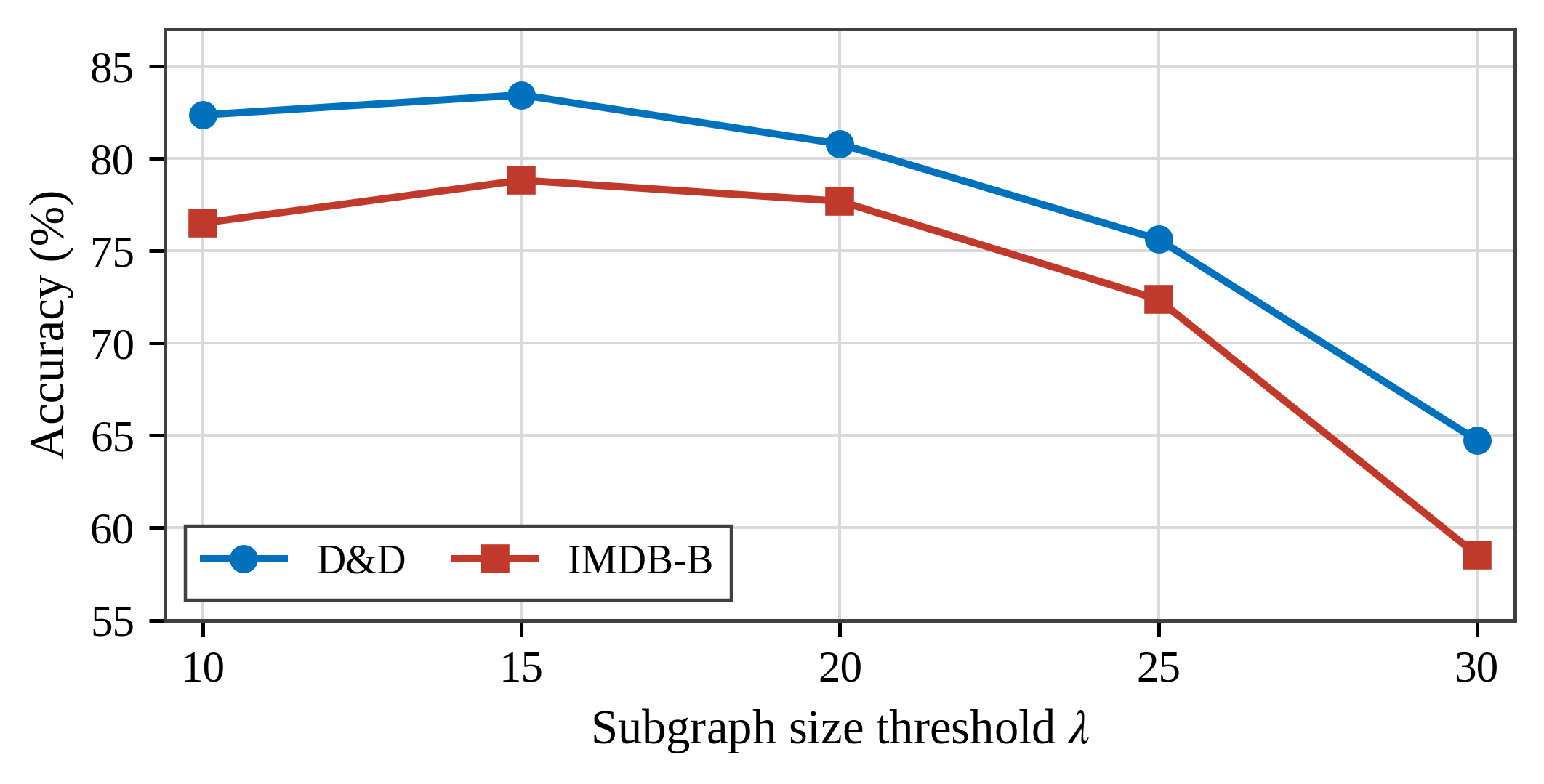}
    \caption{Performance of \CLGC\ for different subgraph size thresholds $\lambda$ on D\&D and IMDB-B.}
    \label{lambda}
 
\end{figure}

Finally, we study the role of subgraph size in compression views. The results in Figure~\ref{lambda} show that the intermediate values of $\lambda$ provide the best performance, with $\lambda=15$ delivering the highest precision in both D\&D (83.43\%) and IMDB-B (78.81\%). When $\lambda$ is too small (e.g., 10), the model captures less structural information and the performance decreases slightly. As $\lambda$ increases further, the compressed graphs become too small, leading to severe information loss. This effect is particularly evident when $\lambda=25$ or $\lambda=30$, where performance drops sharply to 75.62\% and 64.75\% on D\&D, and 72.36\% and 58.54\% on IMDB-B, respectively. These results confirm that, while graph compression is beneficial, excessive compression caused by large subgraph sizes removes important structural patterns and harms classification accuracy.


\section{Conclusion}
In this paper, we propose Structure-aware Compression for Graph Contrastive Learning (\CLGC), a framework that employs $k$-core and $k$-truss decompositions to generate structure-aware graph views for contrastive learning. By constructing compressed graphs that capture both local cohesiveness and higher-order global structures, \CLGC\ provides a principled alternative to random perturbation–based augmentations. Extensive experiments on six benchmark datasets demonstrated that our method outperforms state-of-the-art pooling and contrastive baselines on five of the six datasets, confirming the effectiveness of compression-driven view generation. Ablation studies further highlighted the complementary benefits of $k$-core and $k$-truss decompositions and the importance of balancing supervised and contrastive objectives. Overall, our findings establish graph compression as a powerful strategy for enhancing self-supervised learning in graphs. Future work will extend this framework to dynamic and heterogeneous graphs, as well as explore integration with generative objectives to further improve scalability and representation quality.

\bibliographystyle{ieeetr}
\bibliography{ref}

\end{document}